%% file: main.tex
\documentclass[10pt,twocolumn,letterpaper]{article}

\usepackage[pagenumbers]{wacv} 


\definecolor{wacvblue}{rgb}{0.21,0.49,0.74}
\usepackage[pagebackref,breaklinks,colorlinks,allcolors=wacvblue]{hyperref}

\definecolor{aggregate}{HTML}{69a5a6}
\definecolor{complete}{HTML}{dbbb39}
\definecolor{persist}{HTML}{bc6666}

\usepackage{xcolor}

\usepackage[normalem]{ulem}
\usepackage[most]{tcolorbox}
\def\wacvPaperID{1324} 
\def\confName{WACV}
\def\confYear{2027}

\title{A 3D-Driven Framework for Moving-Camera Perspective-to-360$^\circ$\\Video Generation}

\author{
Michael Wakeham$^1$ \quad Mrinmoy Kumar Das$^1$ \quad Wenqi Wang$^1$\\ Ayush Tewari$^2$ \quad Boqing Gong$^1$ \quad Deepti Ghadiyaram$^1$\\[0.5em]
{\fontsize{10pt}{11pt}\selectfont $^1$Boston University \qquad $^2$University of Cambridge}
}

\begin{document}

\twocolumn[{%
\renewcommand\twocolumn[1][]{#1}%
\maketitle
\centering
\includegraphics[width=0.87\linewidth]{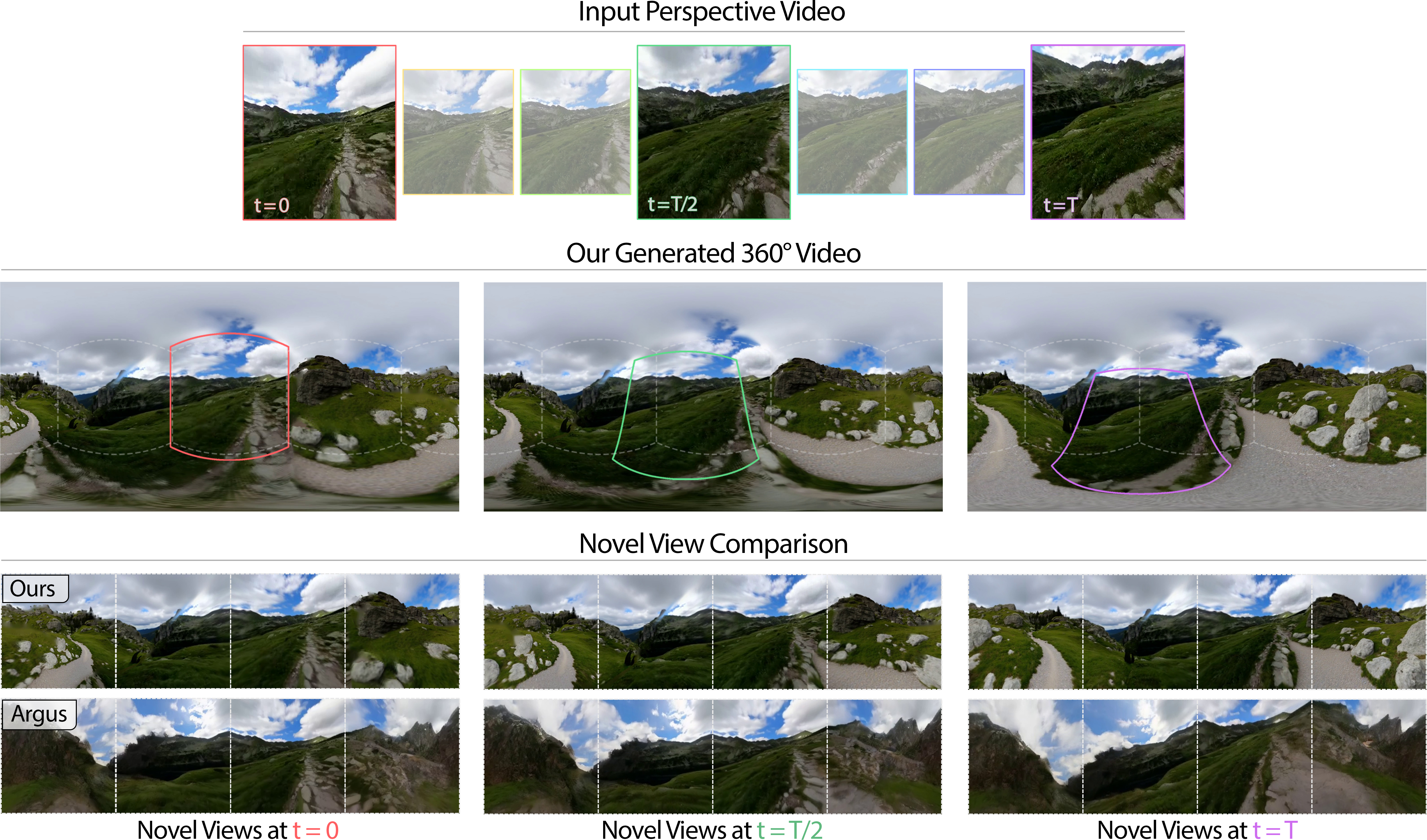}
\captionof{figure}{
\textbf{Moving-camera perspective-to-360$^\circ$ generation.} Given a perspective video of a predominantly static scene  (top), our framework generates a panoramic version that preserves input content while completing unseen panoramic regions (middle). 
The colored outlines show the corresponding input frame projected onto the panoramic output; everything outside this region is generated. Novel views are extracted from the output and displayed on the bottom row, compared with Argus~\cite{argus}. 
Interactive video results and code are available on our project page: 
\href{https://pano3d-video.github.io/}{\nolinkurl{pano3d-video.github.io}}
\vspace{0.9em}}
\label{fig:teaser}
}]

\maketitle

\input{sec/0_abstract}    
\input{sec/1_intro}
\input{sec/2_related_work}
\input{sec/3_method}
\input{sec/4_experiments}
\input{sec/5_conclusion}

{
    \small
    \bibliographystyle{ieeenat_fullname}
    \bibliography{main}
}

\input{sec/X_supplementary}


\end{document}

%% file: sec/0_abstract.tex
\begin{abstract}

Turning a moving-camera perspective video into a 360$^\circ$ video means generating content outside the input frame under several constraints: observations revealed over time must be preserved, unseen regions plausibly completed, and the resulting scene kept consistent as the camera viewpoint changes. Existing methods assign these different requirements to a single conditional video model, requiring them to be jointly learned from panoramic video data. We instead present a modular, 3D-driven framework that distributes these three responsibilities across specialized components, all connected through a shared scene representation. Input observations are reconstructed explicitly in 3D, while a pretrained panoramic image model completes unobserved content from a selected panoramic render. This generated content is lifted into the shared 3D scene, where it remains spatially anchored across camera viewpoints. Without requiring any additional panoramic-video training, our framework achieves state-of-the-art input-view fidelity, perceptual quality, and geometric consistency on a static-scene subset of the Argus evaluation set. These results support our decomposed, 3D-driven formulation as a promising alternative to direct video generation in this setting. 


\end{abstract}

%% file: sec/1_intro.tex
\vspace{-1em}\section{Introduction}
\label{sec:intro}

A captured video's camera framing determines what viewers can observe in a scene and how they experience it. In conventional perspective videos, this framing is entirely fixed: viewers have no control over the field of view, viewing orientation, or camera translation. On the opposite end of the spectrum, 3D environments provide full navigational control, but are considerably harder to capture and recreate. Panoramic video therefore offers a practical middle ground between conventional video and freely navigable 3D. It allows viewers to choose their viewing direction and field of view, while camera translation remains fixed to the captured trajectory.


This makes the moving-camera setting of panoramic video particularly important; translation is solely determined by the capture, yet it also allows viewers to observe the environment from multiple positions. In experiences such as virtual tours or cultural site exploration, the captured trajectory can provide a more immersive experience by revealing scene content that would otherwise not be visible from a static viewpoint. Following this motivation, we study how to generate such moving-camera panoramas from perspective videos of predominantly static scenes.


Existing perspective-to-360$^\circ$ methods~\cite{argus,cubecomposer,viewpoint,imagine360,360anything} largely formulate this task as conditional video generation, typically using a per-frame projection of the input frames as conditioning and relying on a video prior to synthesize the missing panoramic content. While straightforward, this formulation requires the single video model to jointly perform three distinct responsibilities:
\begin{enumerate}
    \item \emph{Aggregation} requires the model to collect content observed anywhere in the input video and preserve it throughout the output, remembering its appearance and inferring where it belongs in the panorama once no longer directly available.
    \item \emph{Completion} requires the model to generate plausible content for regions never observed in the input, while also learning panoramic distortion and seamless boundary continuation. 
    \item \emph{Persistence} 
    requires both observed and generated content to retain consistency in texture and geometry as the camera viewpoint changes. 
\end{enumerate}
For video models, learning these responsibilities jointly is further complicated by the high resolution required for panoramic video, making video-model training particularly expensive. The issue is not that models are incapable of solving the task, but that direct video generation assigns a single expensive model to solve multiple responsibilities with different characteristics at the same time.

Motivated by this, we reformulate the task according to the three defined responsibilities, using a modular framework built around explicit 3D. Rather than regenerating panoramic content at every timestep, we use 3D as a shared scene representation that can be queried for observed content and updated with newly generated content. A feed-forward 3D reconstruction model \emph{aggregates} observations across the input, while a pretrained panoramic image model \emph{completes} the unobserved scene in a single frame. Lifting this generated content into the shared 3D representation allows both observed and generated content to \emph{persist} when rendered along the input camera trajectory.

Through this 3D-driven formulation, we provide a framework that explicitly separates aggregation, completion, and persistence without requiring any additional training on panoramic video data. On a static-scene subset of the Argus~\cite{argus} evaluation set, our framework improves upon the strongest prior results by 1.81 dB on input-view PSNR and 20.4\% on VBench~\cite{vbench} Imaging Quality, while reducing MEt3R~\cite{met3r} geometric inconsistency by 21.8\%. Together, these results establish our 3D-driven formulation as an effective alternative to direct video generation for this moving-camera perspective-to-360$^\circ$ setting.


\begin{figure*}[t]
    \centering
    \includegraphics[width=1\linewidth]{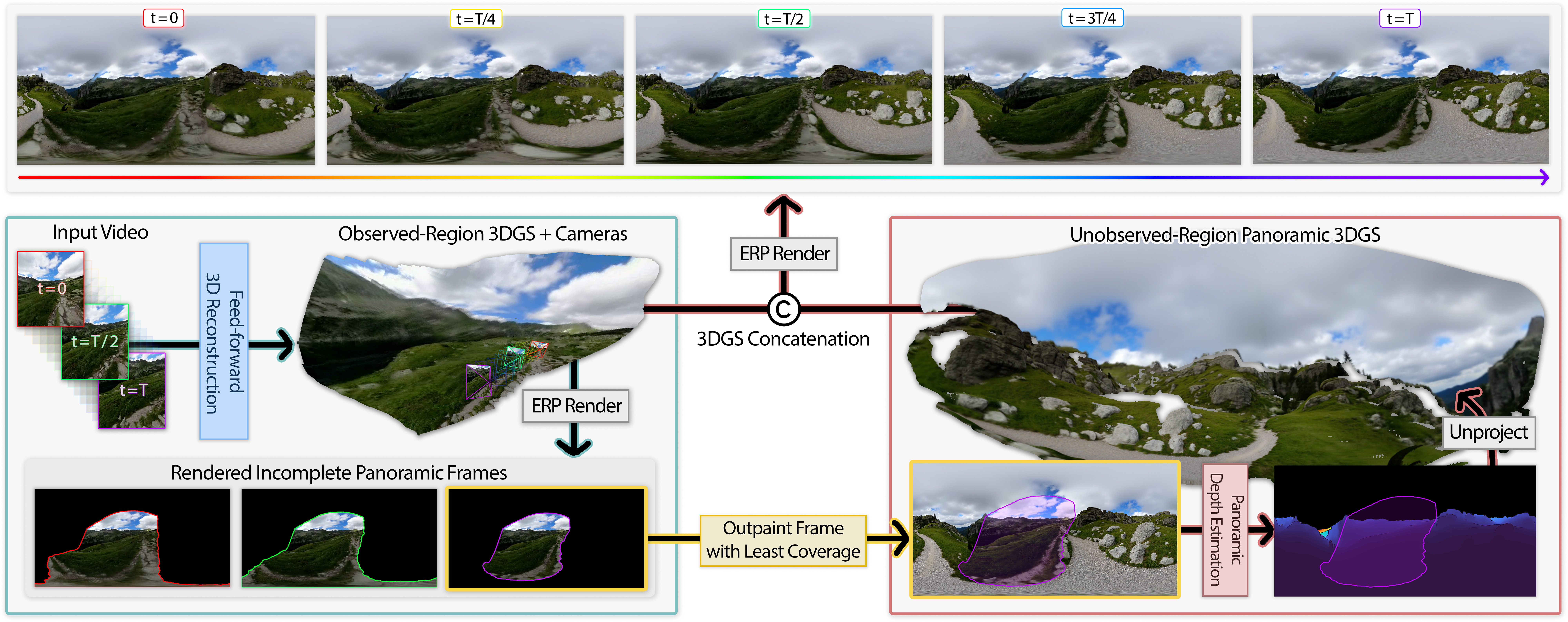}
    \caption{
    \textbf{Framework Overview.}
    A feed-forward 3D reconstruction model \textcolor{aggregate}{\textbf{aggregates}} observations from the input perspective video into an observed-region 3DGS~\cite{3dgs}, which is rendered into incomplete panoramic frames (Sec.~\ref{sec:aggregate}).
    To complete as much unobserved scene content as possible, we select the frame with the least observed coverage and \textcolor{complete}{\textbf{complete}} missing regions using a panoramic image model (Sec.~\ref{sec:complete}). 
    After estimating and aligning its depth, the completed panorama is lifted into an unobserved-region 3DGS and combined with the observed-region 3DGS, allowing the completed scene to \textcolor{persist}{\textbf{persist}} when rendered along the input camera trajectory. (Sec.~\ref{sec:persist}).
    }
    \label{fig:method}
\end{figure*}

%% file: sec/2_related_work.tex
\section{Related Work}
\label{sec:related_work}

\paragraph{Panoramic Generation.}
Panoramic image and video generation look to synthesize complete 360$^\circ$ content while accounting for panoramic distortion and boundary discontinuity. Text-to-panorama methods adapt pretrained diffusion models for panoramic generation~\cite{diffusion360, diffpano, panfusion, dit360, spherediff}, or stitch multiple generated perspective views into a panorama~\cite{cubediff, panofree, jopano}. Image-conditioned methods further complete panoramas from a single observed view through masked generation~\cite{panodiffusion, panodecouple, spnd, dreamcube}. Panoramic video generation extends this task across time, following a similar paradigm of adapting pretrained video models through spherical sampling, panoramic decoding, or interaction between multiple perspective views~\cite{360dvd, dynamicscaler, panowan}. These methods establish strong priors for panoramic content generation, but are limited to image-only generation or text-only conditioning for video. We instead focus on the task of converting perspective video into a panoramic version.

\paragraph{Perspective-to-360$^\circ$ Video Generation.}
More closely related to our task, existing methods generate panoramic videos conditioned on a perspective video input~\cite{vidpanos, imagine360, viewpoint, cubecomposer, argus, 360anything}. Most methods explicitly place or project each input frame into a panoramic representation and use the resulting incomplete video as conditioning. Of these methods, Argus~\cite{argus} and Imagine360~\cite{imagine360} generate directly in the equirectangular space, while others such as CubeComposer~\cite{cubecomposer} and ViewPoint~\cite{viewpoint} operate over a collection of perspective views to better reuse perspective video priors. A more recent work, 360Anything~\cite{360anything}, instead learns the perspective-to-panorama mapping without explicit camera geometry. Despite these differences, all of these approaches rely on a video-generation process to complete missing content and maintain it over time. We instead look to explore a different direction for solving this task through our framework, driven by 3D and not reliant on video priors.

\paragraph{3D Representations for Video Generation.}
Explicit 3D representations provide a natural way to preserve scene content without requiring a video prior to learn this behavior. One such representation, 3D Gaussian Splatting~\cite{3dgs} (3DGS), uses 3D Gaussian primitives for efficient, high-quality 3D modeling. Constructing a 3DGS, however, typically requires camera parameters and scene geometry, which are often unavailable for in-the-wild videos. Recent geometry foundation models~\cite{dust3r, mast3r, vggt, cut3r, pi3} bridge this gap by jointly estimating cameras and scene structure from images, with some methods further predicting a 3DGS scene~\cite{pixelsplat, mvsplat, noposplat, anysplat, yonosplat, worldmirror}. Such advances have been incorporated into generative pipelines by rendering estimated geometry to condition video generation~\cite{camtrol, viewcrafter, trajectorycrafter, gen3c, neoverse}. In comparison, our framework uses explicit 3D for both observed and generated content, rendering the final panoramic video directly without video generation priors. This is closer to perspective-based methods~\cite{srender, wonderworld, wonderjourney} that lift synthesized content into 3D and render as the final output.

%% file: sec/3_method.tex
\section{Method}
\label{method}

Given a perspective video $\mathcal{V} = \{I_t\}^T_{t=1}$ of a predominantly static scene 
captured by a moving camera, our goal is to generate a corresponding equirectangular panoramic (ERP) video along the captured camera trajectory. Our framework follows three stages: aggregation, completion, and persistence, with all stages connected through a shared 3D Gaussian Splatting (3DGS) representation. An overview is given in Figure~\ref{fig:method}.


\subsection{Aggregating Video Observations in 3D}
\label{sec:aggregate}


Our first objective is to aggregate the scene content observed across the input video, rather than treat each perspective frame independently. We achieve this by reconstructing the observed scene into a shared 3D representation, which can subsequently be rendered from each camera position to identify content already supported by the input.

\paragraph{Feed-Forward Scene Reconstruction.} Given the input video $\mathcal{V}$, a feed-forward reconstruction model $\mathcal{F}$ jointly predicts an observed-region 3DGS $\mathcal{G}_\mathrm{obs}$ and a camera $\mathcal{C}_\mathrm{t}$ for each input frame: 
$$(\mathcal{G}_\mathrm{obs}, \{\mathcal{C}_t\}_{t=1}^T) = \mathcal{F}(\mathcal{V}), \quad \mathcal{C}_t = (\mathbf{K}_t, \mathbf{E}_t),$$
here, $\mathbf{K}_t$ and $\mathbf{E}_t$ denote the camera intrinsics and camera-to-world pose at time $t$, respectively. The resulting $\mathcal{G}_\mathrm{obs}$ consolidates content visible across different input frames into a common 3D coordinate system.


\paragraph{Rendering Observed Content.} To query the observed reconstruction, we use an equirectangular renderer $\mathcal{R}_\mathrm{ERP}$, which takes as input the observed 3DGS and pose $\mathbf{E}_t$: 
$$(P_t^\mathrm{obs}, M_t, D_t^\mathrm{obs}) = \mathcal{R}_\mathrm{ERP}(\mathcal{G}_\mathrm{obs}, \mathbf{E}_t)$$ 
The renderer returns an incomplete RGB panorama $P_t^\mathrm{obs}$, a binary visibility mask $M_t$, and rendered depth $D_t^\mathrm{obs}$. The mask takes value 1 at pixels supported by the observed reconstruction and 0 at unsupported pixels. Because the equirectangular projection defines a fixed mapping between output pixels and directions on the viewing sphere, only the camera pose is needed for this rendering.



To establish a consistent ERP coordinate frame, we construct an orientation-stabilized trajectory $\{\widetilde{\mathbf{E}}_t\}_{t=1}^T$ from the estimated trajectory $\{\mathbf{E}_t\}_{t=1}^T$ by preserving each camera translation while replacing its orientation with that of the first frame. We then render $\mathcal{G}_\mathrm{obs}$ at every stabilized pose, producing incomplete panoramas, visibility masks, and depth throughout the trajectory.


\subsection{Completing the Unobserved Panorama}
\label{sec:complete}



The previous stage aggregates scene content revealed across the input video into $\mathcal{G}_\mathrm{obs}$, but cannot supply appearance or geometry for regions that were never observed. The completion stage fills these unsupported regions while retaining the content already represented by $\mathcal{G}_\mathrm{obs}$. Rather than independently generating missing content at every output frame, we select a single candidate panorama for completion and later propagate its generated content through 3D.

\paragraph{Anchor Selection.} 
We refer to this candidate panorama as the anchor. To select it, we first define \textit{observed coverage} at pose $\widetilde{\mathbf{E}}_t$ as the number of supported pixels in the visibility mask, providing a measure for how much scene content is observed at that position. We then choose anchor index $a$ using the camera position with the least observed panoramic coverage:
$$a = \operatorname*{arg\,min}_{t}\lVert M_t \rVert_1$$
Because $M_t$ is binary, $\lVert M_t \rVert_1$ counts the number of pixels supported by the observed reconstruction. Minimizing this quantity selects the panorama containing the largest uncovered region, allowing a single completion to add as much previously unobserved scene content as possible. The selected index $a$ determines the anchor render $P_a^\mathrm{obs}$, its visibility mask $M_a$, and corresponding observed depth $D_a^\mathrm{obs}$.


\paragraph{Masked Panoramic Completion.} For panoramic completion, we use $M_a$ as a keep mask, which preserves existing observed content and uses it to condition the generation of unobserved regions. Given the observed anchor frame, its keep mask, and a generated text prompt $c_\mathrm{text}$, the panoramic completion model $\mathcal{O}$ produces the completed frame $\hat{P}_a$:
$$\hat{P}_a = \mathcal{O}(P_a^\mathrm{obs}, M_a, c_\mathrm{text})$$
We formulate panoramic completion as masked editing of the rendered anchor. Following an inversion-based editing procedure~\cite{personalizeanything}, patches within the keep mask are inverted into the model's latent space and inserted during denoising to preserve masked content. Unmasked patches are generated under $c_\mathrm{text}$ and attend to these inserted latents, producing a completed panorama that retains the observed reconstruction while synthesizing previously unseen regions. 

From this process, we now obtain a completed 2D panorama for a single camera pose; however, its generated content is not yet represented in 3D and cannot be propagated into other camera viewpoints.

\begin{table*}[t]
    \centering
    \renewcommand{\arraystretch}{1.12}
    \resizebox{\textwidth}{!}{%
    \begin{tabular}{lcccccccccc}
        \toprule
        & \multicolumn{5}{c}{Real Camera Trajectory} & \multicolumn{5}{c}{Simulated Camera Trajectory} \\
        \cmidrule(lr){2-6} \cmidrule(lr){7-11}
        Method
        & PSNR $\uparrow$ & LPIPS $\downarrow$ & Imag. $\uparrow$ & Aes. $\uparrow$ & Motion $\uparrow$
        & PSNR $\uparrow$ & LPIPS $\downarrow$ & Imag. $\uparrow$ & Aes. $\uparrow$ & Motion $\uparrow$ \\
        \midrule
        Imagine360~\cite{imagine360}
        & 13.7838 & 0.5771 & 0.4977 & 0.4825 & 0.9706
        & 13.1579 & 0.5943 & 0.5065 & 0.4835 & 0.9681 \\
        ViewPoint~\cite{viewpoint}
        & 15.0143 & 0.5866 & 0.3495 & 0.4495 & 0.9841
        & 13.4829 & 0.6033 & 0.3528 & 0.4554 & 0.9823 \\
        CubeComposer~\cite{cubecomposer}
        & 15.5147 & 0.4918 & 0.4837 & 0.4878 & 0.9840
        & 16.3865 & 0.4764 & 0.4785 & 0.4876 & 0.9837 \\
        Argus~\cite{argus}
        & 16.3249 & 0.4890 & 0.4439 & 0.4880 & 0.9826
        & 16.2471 & 0.4885 & 0.4399 & 0.4720 & 0.9785 \\
        \midrule
        \textsc{Ours}
        & \underline{18.3078} & \underline{0.4325}
        & \textbf{0.6104} & \underline{0.4936} & \underline{0.9872}
        & \underline{17.8771} & \underline{0.4437}
        & \textbf{0.5983} & \underline{0.5020} & \underline{0.9869} \\
        \textsc{Ours + Fill}
        & \textbf{18.3235} & \textbf{0.4324}
        & \underline{0.5730} & \textbf{0.5134} & \textbf{0.9881}
        & \textbf{17.8905} & \textbf{0.4432}
        & \underline{0.5645} & \textbf{0.5173} & \textbf{0.9876} \\

        \bottomrule
    \end{tabular}%
    }
    \caption{
    Quantitative comparison against state-of-the-art perspective-to-$360^\circ$ methods on a static-scene subset of the Argus evaluation set.
    We report PSNR and LPIPS~\cite{lpips} for input-view fidelity, as well as VBench~\cite{vbench} Imaging Quality (Imag.), Aesthetic Quality (Aes.), and Motion Smoothness (Motion) for perceptual video quality.
    We additionally report a post-processed variant of our method (\textsc{Ours + Fill}) that applies lightweight classical inpainting to low-opacity regions.
    }
    \label{tab:main_table}
\end{table*}

\subsection{Persisting and Rendering Content in 3D}
\label{sec:persist}


The final stage makes the completed content persist across the camera trajectory by incorporating it into the shared 3D scene. We estimate its geometry, align it with the observed reconstruction, and lift only the newly generated regions into 3D. Rendering the unified scene propagates both observed and generated content across viewpoints while maintaining coherent structure and appearance.

\paragraph{Panoramic Depth.} To lift this generated content into 3D, we first estimate geometry for the completed anchor using a pretrained panoramic depth estimator $\mathcal{D}$, producing $D_a^\mathrm{pred}=\mathcal{D}(\hat{P}_a)$. Because monocular panoramic depth is scale-invariant, $D_a^\mathrm{pred}$ is not initially expressed at the same scale as the observed reconstruction. To resolve this, the overlap between predicted depth $D_a^\mathrm{pred}$ and rendered depth $D_a^\mathrm{obs}$ is used as a reference for calculating scale alignment:
$$s_a=\operatorname*{median}_{M_a=1}\left(\frac{D_a^{\mathrm{obs}}}{D_a^{\mathrm{pred}}}\right)$$
The aligned depth is then given by $D_a^\mathrm{align} = s_aD_a^\mathrm{pred}$, and can now be used for lifting generated content into 3D.


\paragraph{Lifting Generated Content.} From the aligned depth, we then lift only the newly generated pixels of the completed panorama into the shared coordinate system:
$$\mathcal{G}_{\mathrm{unobs}}=\mathcal{U}(\hat{P}_a,D_a^{\mathrm{align}},1-M_a,\widetilde{\mathbf{E}}_a)$$
Here, $\mathcal{U}$ denotes the function for ERP unprojection, and $1-M_a$ is the complementary visibility mask which selects pixels not supported by the original reconstruction. For each selected pixel, we construct its spherical camera ray, place a 3D Gaussian along that ray according to the aligned depth, and transform this Gaussian into world coordinates using the stabilized anchor pose $\widetilde{\mathbf{E}}_a$. The corresponding RGB pixel initializes the Gaussian color, while aligned depth determines its scale and ERP ray direction determines its orientation~\cite{dreamcube, wonderworld}. Pixels with invalid depth or sharp depth discontinuities are excluded, and the resulting splats are inserted directly without optimization.


\paragraph{Scene Composition and Rendering.} This process produces an unobserved-region 3DGS $\mathcal{G}_\mathrm{unobs}$ containing only newly generated scene content. Because its depth has been aligned and its Gaussians have been transformed into the shared world coordinate frame, it can be concatenated directly with the observed reconstruction:
$\mathcal{G}_{\mathrm{full}}=\mathcal{G}_{\mathrm{obs}}\cup\mathcal{G}_{\mathrm{unobs}}$. The observed-region 3DGS remains responsible for content supported by the input video, while the unobserved-region 3DGS supplies content generated by the panoramic image prior. 

Finally, we render the completed scene along the orientation-stabilized camera trajectory using the equirectangular renderer $\mathcal{R}_\mathrm{ERP}$:
$$(P_t^\mathrm{out}, M_t^\mathrm{out}, D_t^\mathrm{out}) = \mathcal{R}_\mathrm{ERP}(\mathcal{G}_\mathrm{full}, \widetilde{\mathbf{E}}_t), \quad t= 1,...,T$$
The RGB sequence $\{P_t^\mathrm{out}\}^T_{t=1}$ forms the final panoramic video, and because the output is produced through rendering, the camera trajectory can also be interpolated to obtain any desired output frame rate. We also create an optional filled variant, reported in our experiments, where residual holes indicated by $1-M_t^\mathrm{out}$ are filled using a lightweight inpainting post-process~\cite{telea}.

%% file: sec/4_experiments.tex
\section{Experiments}
\label{experiments}

\begin{figure*}[!tp]
    \centering

    \begin{subfigure}{0.96\linewidth}
        \centering
        \includegraphics[width=\linewidth]{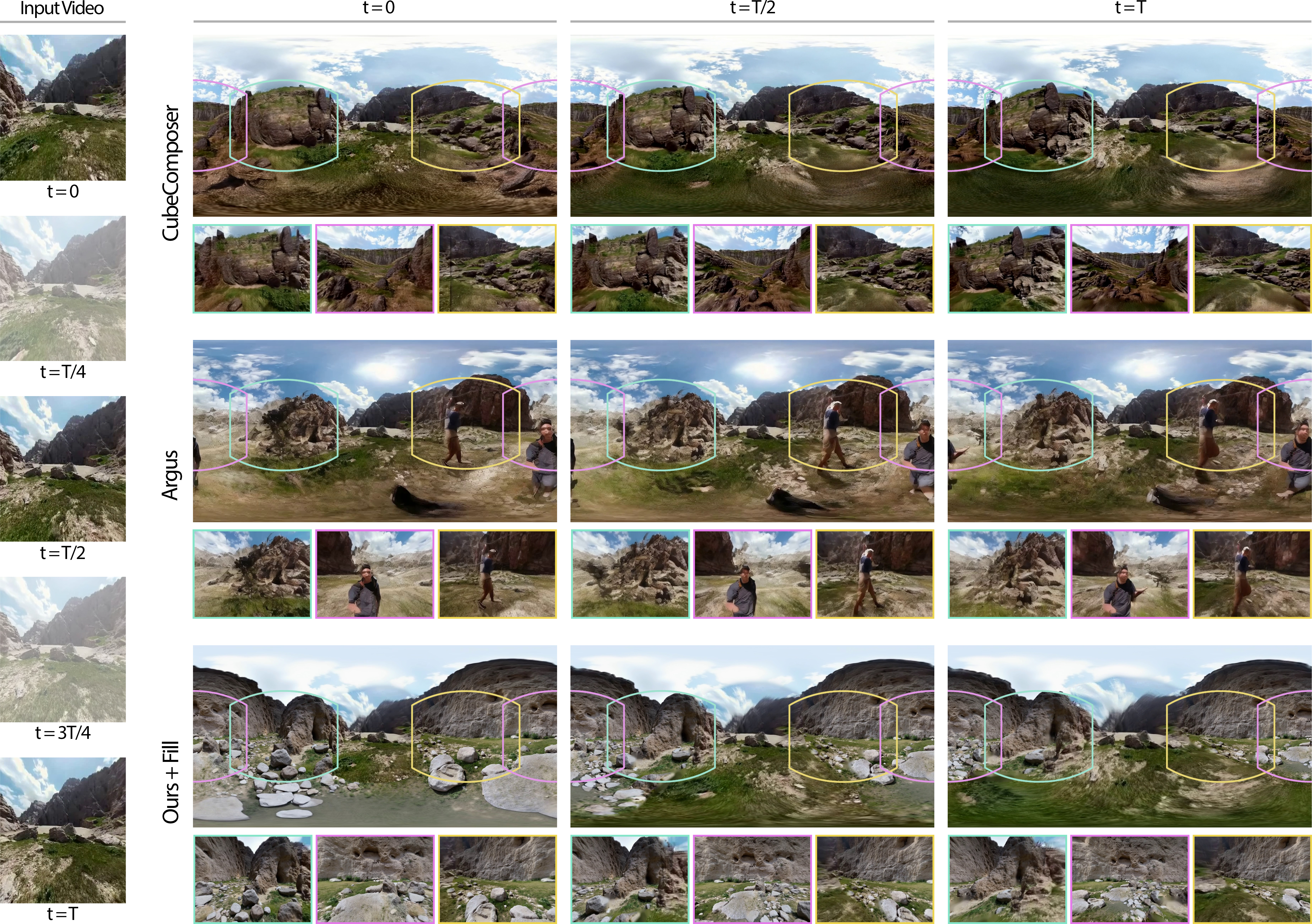}
        \caption{Qualitative comparison with novel views. 
        }
        \label{fig:qualitative-novel}
    \end{subfigure}

    \vspace{0.6em}

    \begin{subfigure}{0.96\linewidth}
        \centering
        \includegraphics[width=\linewidth]{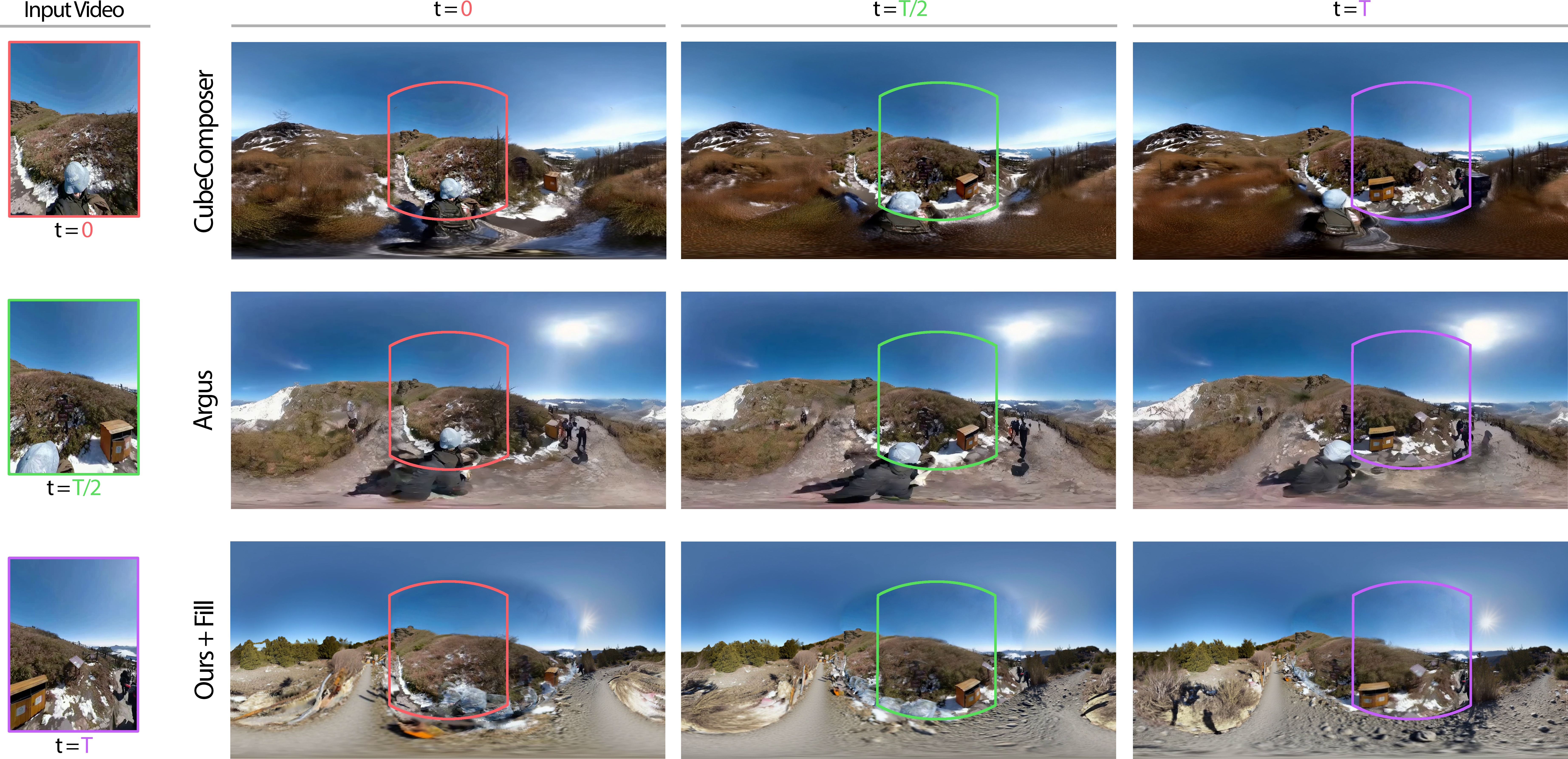}
        \caption{Consistency beyond the projected input.}
        \label{fig:qualitative-consistency}
    \end{subfigure}

    \caption{
    \textbf{Qualitative comparisons.}
    \textbf{(a)} Given the input video (left), we show panoramic outputs at three timesteps. Colored outlines mark novel viewing directions, with the corresponding perspective views below each panorama.
    \textbf{(b)} Colored outlines mark regions explicitly projected from the corresponding input frames. We compare the consistency of the remaining panoramic content across three timesteps. 
    }
    \label{fig:qualitative}
\end{figure*}

We evaluate our framework on a subset of predominantly static scenes from the Argus~\cite{argus} evaluation set, using both real and simulated camera trajectories. The evaluation compares against state-of-the-art perspective-to-360$^\circ$ methods through input-view fidelity, perceptual video quality, and geometric consistency. We further provide qualitative comparisons and analyze our framework design choices, scene coverage, and runtime.

\subsection{Experimental Setup}

\paragraph{Implementation Details.}
We employ HunyuanWorld-Mirror~\cite{worldmirror} as the feed-forward reconstruction model, DiT360~\cite{dit360} as the panoramic image completion model, and PaGeR~\cite{pager} as the panoramic monocular depth estimator. HunyuanWorld-Mirror predictions are used directly without camera priors or splat optimization. For panoramic completion, we use DiT360's inversion-based outpainting, with 50 sampling steps, insertion time of 10, and guidance scale of 2.5. We additionally apply a source-region key scale of 1.05 to strengthen the influence of observed content during denoising. We use Qwen3-VL~\cite{qwen3vl} to generate text prompts, and implement an icosahedron-based panoramic renderer to produce $1024\times2048$ ERP outputs. We report both the direct rendered output, denoted \textsc{Ours}, as well as an optional post-processed variant, \textsc{Ours + Fill}, which applies Telea inpainting~\cite{telea} to low-opacity regions.  All experiments are conducted on a single NVIDIA L40S GPU.

\paragraph{Evaluation Data.}
We construct our evaluation set from the test set used in Argus~\cite{argus}, retaining both its real and simulated camera settings. Following the Argus implementation, we extract perspective inputs at a width of 480 pixels, with the height determined by the corresponding field of view. Because our method targets predominantly static scenes, we filter a subset by removing clips with substantial scene dynamics, severe camera obstruction, or unusably low input resolution. We still retain clips containing minor localized dynamics, such as a small number of moving pedestrians or flowing water, as long as the dominant scene geometry remains static. This results in 35 scenes, each evaluated under both camera-trajectory settings.

\begin{table}[t]
    \centering
    \renewcommand{\arraystretch}{0.96}
    \setlength{\tabcolsep}{5pt}
    \begin{tabular}{lccc}
        \toprule
        & \multicolumn{3}{c}{MEt3R $\downarrow$} \\
        \cmidrule(lr){2-4}
        Method & 5 FPS & 2 FPS & 1 FPS \\
        \midrule
        Imagine360~\cite{imagine360} & 0.090 & 0.146 & 0.200 \\
        ViewPoint~\cite{viewpoint} & 0.067 & 0.111 & 0.147 \\
        CubeComposer~\cite{cubecomposer} & 0.061 & 0.107 & 0.148 \\
        Argus~\cite{argus} & 0.077 & 0.128 & 0.174 \\
        \midrule
        \textsc{Ours} & 0.043 & 0.080 & 0.124 \\
        \textsc{Ours + Fill} & \textbf{0.040} & \textbf{0.075} & \textbf{0.115} \\
        \bottomrule
    \end{tabular}
    \caption{
    Geometric consistency of cubemap faces extracted from generated panoramic videos. We compute MEt3R~\cite{met3r} between consecutive sampled frames, where lower sampling rate compares frames farther apart in time.
    }
    \label{tab:met3r}
\end{table}

\paragraph{Metrics.}
We mainly follow the Argus evaluation protocol, with a modified input-view fidelity evaluation detailed in the supplementary. Input-view fidelity is measured using PSNR and LPIPS~\cite{lpips} between the ground-truth perspective crop and the corresponding view extracted from each generated panorama. Perceptual and temporal video quality are measured using three VBench~\cite{vbench} dimensions: Imaging Quality, Aesthetic Quality, and Motion Smoothness. Each VBench metric is calculated over the left, front, right, and back cubemap-face videos extracted from the panoramic output. We use MEt3R~\cite{met3r} to measure geometric consistency between frames on the same four cubemap faces, calculated between consecutive frames sampled at 1, 2, and 5 FPS. For fairness, all methods are evaluated at matched resolutions, with higher-resolution outputs downsampled.

\paragraph{Baselines.}
We compare against four publicly available state-of-the-art perspective-to-360$^\circ$ methods: Imagine360~\cite{imagine360}, ViewPoint~\cite{viewpoint}, CubeComposer~\cite{cubecomposer}, and Argus~\cite{argus}. We preserve the camera-conditioning procedure used by each method rather than imposing a shared projection scheme. Argus and CubeComposer project the input using camera poses estimated by MASt3R~\cite{mast3r}, ViewPoint places the input perspective video on the front cubemap face, and Imagine360 uses its own estimation of pitch and field of view. We do not include 360Anything~\cite{360anything} as no public code/weights are available.

\subsection{Results}
\paragraph{Quantitative Comparisons.}
Table~\ref{tab:main_table} reports input-view fidelity and VBench results for real and simulated camera trajectories. Both variants of our method consistently achieve the strongest PSNR and LPIPS, demonstrating improved agreement between our reconstructed input views and the ground truth cropped views. Our direct render also achieves the highest Imaging Quality by a considerable margin, and our fill variant achieves the highest Aesthetic Quality. Motion Smoothness is nearly saturated across all methods and is therefore less informative.

Our two variants exhibit a clear tradeoff. \textsc{Ours} retains the sharper appearance of the direct 3D render and performs best in Imaging Quality, but worse in Aesthetic Quality. On the other hand, \textsc{Ours + Fill} fills residual holes from the render and performs worse than \textsc{Ours} on Imaging Quality, but achieves the best score for Aesthetic Quality. 

We also measure geometric consistency, reported in Table~\ref{tab:met3r}. Both of our variants achieve the lowest MEt3R error, with our advantage holding at every sampling rate. This supports the benefit of rendering each output from an explicitly shared 3D scene, rather than relying on the learned temporal consistency from a video model.

\begin{table}[t]
    \centering
    \renewcommand{\arraystretch}{1.08}
    \setlength{\tabcolsep}{2pt}
    \resizebox{\columnwidth}{!}{%
    \begin{tabular}{lcccc}
        \toprule
        Variant
        & PSNR $\uparrow$
        & LPIPS $\downarrow$
        & VBench $\uparrow$
        & MEt3R $\downarrow$ \\
        \midrule
        \multicolumn{5}{l}{\textit{Framework ablations (full evaluation set)}} \\
        Panorama-Only 3D
            & 14.382 & 0.538 & 0.695 & 0.083 \\
        Unmasked 3D Lifting
            & 16.979 & 0.483 & 0.689 & 0.086 \\
        Single-Frame Cond.
            & 16.890 & 0.466 & \textbf{0.700} & 0.086 \\
        \textsc{Ours}
            & \textbf{18.092} & \textbf{0.438} & 0.696 & \textbf{0.082} \\
        \midrule
        \multicolumn{5}{l}{\textit{Persistence ablation (random 15 scene subset)}} \\
        Per-Frame Completion
            & 15.793 & 0.473 & \textbf{0.717} & 0.216 \\
        \textsc{Ours}
            & \textbf{17.941} & \textbf{0.435} & 0.698 & \textbf{0.074} \\
        \bottomrule
    \end{tabular}%
    }
    \caption{
    Framework and persistence ablations, where VBench averages the three reported dimensions, while MEt3R averages the three sampling rates. 
    \textit{Panorama-Only 3D} discards the input reconstruction, keeping only the 3D-lifted panorama.
    \textit{Unmasked 3D Lifting} lifts the full panorama and fuses it without input-observation masking, allowing generated content to overlap existing geometry.
    \textit{Single-Frame Cond.} conditions the outpainting on a single input frame rather than the rendered full-video reconstruction.
    \textit{Per-Frame Completion} separately completes every incomplete rendered panoramic frame without any 3D persistence.
    }
    \label{tab:ablation}
\end{table}

\paragraph{Qualitative Comparisons.}
Figure~\ref{fig:qualitative-novel} compares our method with Argus and CubeComposer using novel perspective views extracted from the generated panoramas. These views make the visual quality of the panorama easier to inspect. Our method produces sharper details and clearer scene content, particularly noticeable in the foreground rocks, ground textures, and background valley, while the baselines appear blurred and smeared.

Figure~\ref{fig:qualitative-consistency} instead examines consistency beyond the projected input. The colored outlines mark the regions explicitly provided as conditioning, so we focus on the remaining panoramic content. Across different timesteps, our method maintains a more coherent scene appearance, whereas the baselines exhibit greater variation in texture and structure.

Additional qualitative results and interactive video comparisons are provided on our project page.


\subsection{Ablation and Analysis}
\label{ablation_and_analysis}

\paragraph{Framework Ablations.}
We hold the pretrained components fixed and ablate the framework-level choices in the upper panel of Table~\ref{tab:ablation}, with each ablation defined in the caption. \textit{Panorama-Only 3D} causes the largest reduction in input-view fidelity, demonstrating that a lifted single-view panorama cannot replace observations aggregated across the input video. \textit{Single-Frame Cond.} and \textit{Unmasked 3D Lifting} also reduce fidelity, showing the respective benefits of full-video conditioning and separating generated geometry from input-supported content. VBench and MEt3R remain similar across variants, but for different reasons: perceptual quality is largely governed by the panoramic completion model, while every variant still obtains geometric consistency by rendering from explicit 3D.

\begin{table}[t]
    \centering
    \renewcommand{\arraystretch}{1.12}
    \resizebox{\columnwidth}{!}{%
    \begin{tabular}{lcc}
        \toprule
        Method / Stage & Resolution & Runtime (s) \\
        \midrule
        \multicolumn{3}{l}{\textit{Runtime comparison between methods}} \\
        Imagine360~\cite{imagine360} & 1K & 386.13 \\
        ViewPoint~\cite{viewpoint} & 1K & 82.92 \\
        CubeComposer~\cite{cubecomposer} & 2K & 582.26 \\
        Argus~\cite{argus} & 1K & 139.59 \\
        \textsc{Ours} & 2K & 267.86 \\
        \textsc{Ours + Fill} & 2K & 287.35 \\
        \midrule
        \multicolumn{3}{l}{\textit{Runtime breakdown of our method}} \\
        3D reconstruction & --- & 0.98 \\
        Panoramic outpainting & 2K & 245.09 \\
        Panoramic depth estimation & 2K & 2.01 \\
        Panorama rendering & 2K & 9.72 \\
        Other overhead & --- & 10.06 \\
        Classical inpainting (optional) & 2K & 19.49 \\
        \bottomrule
    \end{tabular}%
    }
    \caption{
    Runtime comparison and breakdown for 5-second videos on a single NVIDIA L40S GPU. The comparison reports total runtime for generating 25 frames at 5 FPS.
    }
    \label{tab:runtime}
\end{table}

\paragraph{Persistence Ablation.} The lower panel of Table~\ref{tab:ablation} evaluates \textit{Per-Frame Completion}, where we separately apply the panoramic image completion model to the rendered incomplete panorama at each timestep. This setup preserves the full-video aggregation but removes the lifting and propagation of generated content through the shared 3D scene. Because of computational cost, we evaluate this ablation on only fifteen randomly sampled scenes. We test on both the real and simulated camera trajectory at full 5 FPS output, and fix all hyperparameters, text prompts, and seeds. 

The results show that \textit{Per-Frame Completion} achieves a slightly higher average VBench score, but noticeably reduces input-view fidelity and substantially increases MEt3R error. This validates our separation of completion and persistence, as independently generating high-quality frames is insufficient without a shared representation that persists content over time.

\paragraph{Scene Coverage.} We measure how much of the final video is directly supported by rendered 3D content using a scene coverage metric:
$$\operatorname{Coverage} = \frac{1}{THW}\sum^T_{t=1}\lVert M_t^\mathrm{out} \rVert_1$$
where $M_t^\mathrm{out}$ is the binary output render mask, with 1 indicating a pixel supported by the 3D scene. We calculate this metric for each video and average it across the evaluation set. Our framework achieves 96.385\% scene coverage across real and simulated camera trajectories, leaving only 3.615\% of output pixels for optional filling. This shows that, although completion is performed at only one anchor, the unified 3D scene still provides dense render support throughout the camera trajectory.

\paragraph{Runtime.}
Table~\ref{tab:runtime} compares inference runtime between our method and the baselines, including a breakdown of our method's runtime by component. Our method produces 2K resolution output faster than Imagine360 and CubeComposer, although it is slower than the 1K outputs of ViewPoint and Argus. Panoramic outpainting accounts for most of our runtime and remains the primary computational bottleneck. Since panoramic rendering is comparatively inexpensive, higher output frame rates incur relatively little additional cost.

%% file: sec/5_conclusion.tex
\section{Conclusion}
\label{conclusion}
In this work, we introduced a 3D-driven formulation for moving-camera perspective-to-360$^\circ$ generation, built around three responsibilities: aggregation, completion, and persistence. Rather than assigning these responsibilities jointly to a single video model, our framework separates them across specialized components connected through a shared 3D scene representation. Without requiring additional panoramic video training, this design improves input-view fidelity, visual quality, and geometric consistency over existing state-of-the-art methods. These results support explicit 3D as an effective alternative to direct video generation in this setting.

\paragraph{Limitations.} Our framework is designed for predominantly static scenes and is not capable of modeling independently moving objects, which may be incorrectly incorporated into the shared 3D representation (see Figure~\ref{fig:qualitative-consistency}). In addition, because generated geometry comes from a single anchor, large translations may reveal unsupported surfaces; multiple anchors or iterative completion could address this limitation. Finally, the performance and efficiency of our framework depend on the models used for reconstruction, completion, and depth estimation; the modular design, however, allows these components to be replaced as stronger models become available.

%% file: sec/X_supplementary.tex
\clearpage

\setcounter{section}{0}
\renewcommand{\thesection}{\Alph{section}}
\renewcommand{\theHsection}{supp.\Alph{section}}

\setcounter{figure}{0}
\renewcommand{\thefigure}{S\arabic{figure}}

\setcounter{table}{0}
\renewcommand{\thetable}{S\arabic{table}}

\setcounter{equation}{0}
\renewcommand{\theequation}{S\arabic{equation}}

\renewcommand{\theHfigure}{supp.\arabic{figure}}
\renewcommand{\theHtable}{supp.\arabic{table}}

\maketitlesupplementary

\section{More Qualitative Results}

Additional qualitative comparisons are provided in the project page gallery. The gallery presents the input perspective videos alongside outputs from Imagine360~\cite{imagine360}, ViewPoint~\cite{viewpoint}, CubeComposer~\cite{cubecomposer}, Argus~\cite{argus}, and \textsc{Ours + Fill}. Each ERP video can be opened in an interactive $360^\circ$ viewer, allowing viewers to pan and zoom when examining spatial and temporal artifacts that are difficult to assess from static figures. The gallery also provides a side-by-side visualization of the framework and persistence ablations for an example scene. Ablation outputs are shown without optional filling, consistent with the protocol described in Section~\ref{sec:ablation_details}. 

\begin{table}[t]
\centering
\small
\begin{tabular}{@{}p{0.07\linewidth}p{0.89\linewidth}@{}}
\toprule
\textbf{ID} & \textbf{Caption} \\
\midrule

0006 & ``This is a panorama image. The photo shows a wooded roadside scene with distant tall trees, a curving asphalt road, leaf-covered ground, and a partly cloudy sky.'' \\

0007 & ``This is a panorama image. The photo shows a vast rocky canyon with steep walls, a shallow stream, stones, and an opening of sky above.'' \\

0008 & ``This is a panorama image. The photo shows a paved pathway between a building and a hillside in soft sunlight with a bright cloudy sky.'' \\

0009 & ``This is a panorama image. The photo shows an open campus courtyard with a paved pathway, green lawn, distant stone academic buildings, and a cloudy sky.'' \\

0011 & ``This is a panorama image. The photo shows a rural scene with a paved path and green fields under a clear sky.'' \\

0015 & ``This is a panorama image. The photo shows a dark forest path winding through distant dense trees, with fallen leaves on the ground and sunlight filtering through the canopy.'' \\

0017 & ``This is a panorama image. The photo shows a rural hillside landscape with distant scattered trees and open greenery under a clear sky.'' \\

0018 & ``This is a panorama image. The photo shows a tree-lined path with distant tall evergreens, cherry blossoms, asphalt pavement, and traditional structures under a partly cloudy sky.'' \\

0021 & ``This is a panorama image. The photo shows a vast canyon landscape with a winding river below, rocky cliffs, and a clear bright sky.'' \\

0023 & ``This is a panorama image. The photo shows a garden path with stone steps, dense greenery, and many small flowers.'' \\

0025 & ``This is a panorama image. The photo shows a spacious outdoor area with pathways, green fields, scattered umbrellas, distant hills, and a city skyline under a clear sky.'' \\

0026 & ``This is a panorama image. The photo shows an open rural landscape with grass, a building on the left, a tree in the foreground, with hills and trees in the distance.'' \\

0027 & ``This is a panorama image. The photo shows a large sunny park pathway lined with green lawns, leafy trees, benches, and soft shadows under a partly cloudy sky.'' \\

0030 & ``This is a panorama image. The photo shows a wide urban street with parked cars, multi-story buildings, sidewalks, and a bright daytime sky.'' \\

0033 & ``This is a panorama image. The photo shows a forest path winding through distant tall trees, with earthy ground, ferns, and dense green vegetation.'' \\

0036 & ``This is a panorama image. The photo shows a wet rural road running through grassy fields, with fences, distant small buildings, hills, and bright clouds in a blue sky.'' \\

0037 & ``This is a panorama image. The photo shows a wide dirt forest trail surrounded by dark dense greenery, distant tall trees, ferns, and broad leaves.'' \\

0043 & ``This is a panorama image. The photo shows a spacious campus courtyard with brick buildings, stone pathways, and a clear bright sky.'' \\

\bottomrule
\end{tabular}
\caption{generated captions for the evaluation scenes (Part 1).}
\label{tab:scene_captions_1}
\end{table}

\begin{table}[t]
\centering
\small
\begin{tabular}{@{}p{0.07\linewidth}p{0.89\linewidth}@{}}
\toprule
\textbf{ID} & \textbf{Caption} \\
\midrule

0044 & ``This is a panorama image. The photo shows a campus pathway beside a brick building, with grass, stone borders, and an overcast sky.'' \\

0046 & ``This is a panorama image. The photo shows a mountain trail winding through sparse vegetation, snow patches, rocky terrain, and a clear sky.'' \\

0049 & ``This is a panorama image. The photo shows a park with blooming trees, a calm pond, green grass, and a clear blue sky.'' \\

0051 & ``This is a panorama image. The photo shows a wide urban park beside a river, with grass, pathways, distant skyscrapers, and a curved bridge under a clear sky.'' \\

0054 & ``This is a panorama image. The photo shows a mountain trail with stone steps winding through grassy alpine terrain, scattered rocks, distant peaks, and a partly cloudy sky.'' \\

0056 & ``This is a panorama image. The photo shows a wide park pathway leading toward a stone building, with green grass, white flowers, and an overcast sky.'' \\

0058 & ``This is a panorama image. The photo shows a calm waterside scene with a tree-lined shore, a modern building, gravel and grass along the water, and an overcast sky.'' \\

0060 & ``This is a panorama image. The photo shows the outside of a prominent institutional building with a central entrance, an asphalt road, lampposts, and a partly cloudy sky.'' \\

0061 & ``This is a panorama image. The photo shows a mountain trail winding through grassy slopes, with stones, mossy rocks, distant peaks, and a wide cloudy sky.'' \\

0063 & ``This is a panorama image. The photo shows a wide forest path through dense grass, distant trees, and patches of sunlight under a clear sky.'' \\

0067 & ``This is a panorama image. The photo shows a vast rocky canyon with steep pale rock walls, a shallow clear stream, stones, and reddish cliff faces.'' \\

0071 & ``This is a panorama image. The photo shows a vast mountainous landscape from a grassy slope, with green terrain, distant ridges, and a dark vibrant blue sky.'' \\

0074 & ``This is a panorama image. The photo shows a rocky canyon landscape with large boulders, a shallow riverbed, steep cliffs, patches of grass, and a partly cloudy sky.'' \\

0076 & ``This is a panorama image. The photo shows a forested hillside beside a winding paved road, with fallen leaves, tall trees, and a cloudy sky.'' \\

0081 & ``This is a panorama image. The photo shows a paved pathway beside a concrete wall, with a wooded area, a small shelter, a riverbank, and a partly cloudy sky.'' \\

0086 & ``This is a panorama image. The photo shows a quiet urban street with parked cars, brick buildings, sidewalks, red brick pavement, and a cloudy sky.'' \\

0093 & ``This is a panorama image. The photo shows a brick residential building facade with windows, black shutters, balconies, a lamppost, and a brick sidewalk.'' \\

\bottomrule
\end{tabular}
\caption{generated captions for the evaluation scenes (Part 2).}
\label{tab:scene_captions_2}
\end{table}

\section{Additional Implementation Details}

We use the same notation and definitions from the main text and provide additional details to choices not previously specified. All components use their publicly released implementation and weights, no optimization or fine-tuning is done to the models or 3D Gaussian Splats, and we use gsplat~\cite{gsplat} for 3DGS rasterization.

\paragraph{Models and input processing.} We instantiate aggregation with HunyuanWorld-Mirror v1~\cite{worldmirror}, completion with the \texttt{FLUX.1-dev} backbone and released DiT360 LoRA~\cite{dit360}, and panoramic depth with the released PaGeR checkpoint~\cite{pager}. Reconstruction is done over 2 FPS sampled frames from the original 25 frame, 5 FPS evaluation videos. The estimated camera trajectory can then be interpolated to any desired output FPS. After obtaining the interpolated camera trajectory, we stabilize it and use that stabilized trajectory throughout the rest of the pipeline.

\paragraph{Coverage computation.}
Candidate anchors are scored at each reconstructed pose from the 2 FPS estimated trajectory. Each coverage score uses a $1024{\times}512$ ERP alpha image assembled from six $256{\times}256$, $90^\circ$ cubemap renders. We do not use the full 2K resolution icosahedron rendering at this stage since anchor selection only requires a relative ranking of coverage. We additionally filter any unusually large 3D Gaussians to prevent isolated or diffuse splats from artificially increasing the observed area. A pixel is considered covered when its filtered alpha exceeds 0.5.

\paragraph{Icosahedron ERP rendering.} Whenever a full-resolution panorama is required, we render the 3DGS through 12 fixed perspective cameras whos optical axes point towards the 12 vertices of an icosahedron. Each camera has a $90^\circ$ field of view, producing overlapping perspective images that jointly cover the viewing sphere. We reproject these images into equirectangular space and combine overlapping RGB and alpha values using squared-cosine angular weights, giving greater weight to views in which a pixel lies closer to the icosahedron vertex. We do not blend depth values, since averaging depth from different views can create incorrect intermediate geometry. Instead, each ERP pixel takes its depth from the camera best viewing its ray direction. We render the perspective views at $1024\times1024$ resolution as a supersampling step before stitching them into a $2048\times1024$ panorama.

\paragraph{Conditioning and masked panoramic completion.} At the selected anchor pose, we construct a $2048\times1024$ panoramic RGB condition using the icosahedron renderer. Alpha values above 0.5 are masked as pixels to preserve within the keep mask. Before VAE inversion, we conservatively downsample the keep mask to the latent patch resolution. This ensures that a latent patch is only preserved when its entire region is observed, preventing the unrendered buffer from incorrectly seeping into the output as an observed region. We use 50 inversion and sampling steps, seed 29, insertion parameter 10, guidance scale 2.5, and source-key scale 1.05.

\paragraph{Depth estimation and alignment.} PaGeR~\cite{pager} automatically selects an indoor or outdoor prediction mode and returns both panoramic depth and sky-validity mask. We filter out unreliable depth near discontinuities by comparing each pixel with its four immediate neighbors. A pixel is rejected when both the absolute and relative depth difference exceed 0.1. Scale alignment uses pixels that are both observed in the keep mask and valid according to PaGeR. We then discard the outer 10\% of the depth-ratio distribution and use the median of the remaining ratios to calculate the scale alignment. Pixels classified as sky are assigned the farthest valid depth so that generated sky content is placed in the background, while rejected pixels from depth discontinuities are not lifted.

\paragraph{3D Gaussian initialization and filtering.} When unprojecting generated content into 3D, a new Gaussian is created for each pixel outside the keep mask using the aligned panoramic depth. Each Gaussian is initialized as a surfel, where its horizontal and vertical scales match the angular area covered by one ERP pixel at the predicted depth~\cite{dreamcube, wonderworld}. The orientation of the surfel is aligned with the viewing ray. The RGB value at the pixel determines the color of the Gaussian, and its opacity is set to 0.99.

\paragraph{Final rendering and optional filling.} Final ERP frames are rendered using the icosahedron renderer at $2048\times1024$ resolution. We recover the final RGB values by removing alpha premultiplication wherever the rendered alpha is at least 0.1; pixels below this threshold are marked as uncovered. Our direct output retains these uncovered regions. For the optional \textsc{Ours + Fill} variant, we apply Telea inpainting independently with a radius of 15 to the uncovered pixels in each frame. This filling is performed entirely in image space after rendering and does not modify the Gaussian scene or original content in each frame.

\section{Evaluation Protocol}

\subsection{Evaluation Data}

\paragraph{Evaluation-set construction.} We construct the evaluation set using perspective extractions from ground truth equirectangular videos in Argus~\cite{argus} evaluation set. The camera trajectories and fields of view are also provided by Argus. We evaluate on both their real and simulated camera trajectories and follow their implementation for input extraction: 25 frames at 5 FPS with a fixed width of 480 pixels. Since the scope of our method is mainly for predominantly static scenes, we do not use all 101 videos from the original Argus test set. We instead filter a subset by excluding clips that contain substantial scene dynamics, severe physical camera obstruction, or unusably low perspective-view resolution. Minor dynamics such as isolated people or moving water are retained. This results in 35 scenes across both real and simulated trajectories, totaling to 70 evaluated cases.

\paragraph{Generating captions.} We use Qwen3-VL-8B-Instruct~\cite{qwen3vl} to generate captions for each video in the filtered subset of Argus~\cite{argus} test set. We feed the extracted perspective video at 1 FPS to the VLM and use the same prompt across both real and simulated camera trajectories. In addition, a prefix is added to the generated caption, ``This is a panorama image. The photo shows'', as DiT360~\cite{dit360} is always trained with this prefix. When using the prompt for other baseline models, we remove the prefix. We provide the prompt used for generation as well as the IDs of each video in the evaluation subset and generated prompts at Table~\ref{tab:scene_captions_1} and Table~\ref{tab:scene_captions_2}. 
\begin{tcolorbox}[
    colback=gray!5,
    colframe=gray!40,
    boxrule=0.5pt,
    arc=2pt,
    left=5pt,right=5pt,top=5pt,bottom=5pt
]
\small
\texttt{
Caption this environment as if it were one wide photograph, not a recap of the footage. Keep it very simple and general: what kind of place it is and what is the basic layout of the scene without directional descriptors. Keep within a 40 word count and start the caption text with 'The photo shows' and then continue. For the entire caption, write the place as fixed at a single instant: the setting itself, not how the view travels, and not things that only read as momentary or in motion.
}
\end{tcolorbox}

\subsection{Baseline Setup}

We use publicly released implementations and checkpoints for Argus~\cite{argus}, CubeComposer~\cite{cubecomposer}, Viewpoint~\cite{viewpoint}, and Imagine360~\cite{imagine360}. All methods receive the same extracted perspective input video and scene caption (if applicable). Their outputs are evaluated over the same 25 timestamps spanning 5 seconds. Imagine360 and CubeComposer generate slightly more than 25 frames, so we pad the inputs and then trim outputs to the first 25 frames. We retain each method's camera-conditioning setup and output resolution; resolution normalization is instead performed inside each metric.

\subsection{Metrics}

\paragraph{Input-view fidelity.} We use PSNR and LPIPS to measure preservation of observed input content. Generated and reference ERPs are first downsampled to the same resolution of $1024\times512$ (if applicable). At every timestep, the ground truth camera rotation and field of view are used to extract corresponding perspective views from the ERP frame. We use the same Argus implementation for extracting these generated views as the ground truth inputs. The resulting images have the same $480\times H$ resolution as the inputs and are compared over the complete frame.

Our protocol for extracting these views for comparison differs from the evaluation used by Argus. Argus evaluates over the union of each ERP-level mask, resulting in one ERP-space mask applied over every ERP frame for calculating PSNR and LPIPS. This formulation for evaluation has two limitations: (1) a temporal union contains all scene content observed at \textit{some} point in the video, but under camera motion or scene dynamics, those directions are not necessarily observed at \textit{all} timesteps. Reconstruction metrics are therefore being evaluated over not only input-reconstructed regions, but also generated regions, where reconstruction metrics do not work well as a form of evaluation. (2) standard LPIPS using a standard VGG backbone is trained on \textit{regular perspective} images with \textit{rectangular shape}. However in this setting, LPIPS is calculated within the \textit{ERP-distorted} image space, and must only be calculated over an \textit{irregularly-shaped} mask. This departs from the unmasked perspective-image input that LPIPS is usually calculated on.

\paragraph{Perceptual quality.} We evaluate VBench~\cite{vbench} Imaging Quality, Aesthetic Quality, and Motion Smoothness following the same Argus protocol. Each ERP video is converted into front, back, left, and right $256\times256$ perspective videos; top and bottom views are excluded. Each dimension is computed independently for each face, and face scores are averaged to obtain a video score.

\paragraph{Geometric consistency.} MEt3R~\cite{met3r} is evaluated on the same four cubemap faces that perceptual quality evaluation uses, also at $256\times256$ resolution. We evaluate at 5, 2, and 1 FPS by selecting the frame nearest to each sampling timestamp and forming pairs between consecutive selected frames. The 5 FPS setting compares adjacent frames from the original video, while 2 and 1 FPS compare frames separated by progressively greater temporal difference. This multi-framerate evaluation distinguishes short-term frame-by-frame smoothness from longer-range geometric persistence.

\paragraph{Scene coverage.} We measure scene coverage using the binary visibility ERP masks produced by our renderer as final output. Pixels with alpha above 0.1 are considered "covered" and set to full opacity. This threshold retains valid but partially transparent support near Gaussian boundaries; an otherwise stricter threshold would incorrectly classify as a hole and then (optionally) inpaint.

\section{Ablation Details}
\label{sec:ablation_details}

\subsection{Framework Ablations} The framework ablations are evaluated on all 35 scenes under both real and simulated trajectories. Unless directly modified by an ablation, all variants use the same configurations as the original full pipeline. We evaluate on direct rendered outputs without the optional filling.

\paragraph{Panorama-Only 3D.} We retain the same reconstruction, anchor selection, panorama completion, depth estimation, and depth alignment as the full method. However, rather than lifting only the unobserved region and merging it with the input reconstruction, we lift the entire completed panorama into a new 3DGS. The original reconstructed 3DGS is then discarded, so both observed and generated scene content are represented solely by the generated panorama's geometry and appearance. This isolates the benefit of retaining the original reconstructed observations, rather than representing the complete scene through a single generated panorama.

\paragraph{Unmasked 3D Fusion.} The full method prevents generated content from being lifted wherever the input reconstruction already provides support. In this variant, that restriction is removed; the complete outpainted panorama is converted into a full 3DGS and fused with the reconstructed 3DGS. As a result, generated regions of the scene can overlap with their corresponding input-reconstructed regions. This tests whether visibility-aware fusion is necessary to protect input-reconstructed geometry and appearance.

\paragraph{Single-Frame Conditioning.} Instead of creating the DiT360 condition from the full-video reconstruction, this variant projects the middle input frame into an ERP at its corresponding stabilized pose. All downstream stages, including depth estimation, masked lifting, and rendering, remain unchanged. The difference is therefore whether panoramic completion can condition on observations gathered over the entire video or only the content visible in one perspective frame.

\subsection{Persistence Ablation.}

\paragraph{Per-Frame Completion.} This variant renders an incomplete stabilized panorama at each output pose and completes all 25 frames independently with the same prompt, seed, and sampling configuration. The completed panoramas are assembled directly into a video, bypassing generated-depth estimation, lifting, and fusion. This tests whether temporal persistence can arise from repeatedly querying the image prior without persisting content through a shared 3D representation.

%% file: main.bib
@String(CVPR= {IEEE Conf. Comput. Vis. Pattern Recog.})

@String(ICCV= {Int. Conf. Comput. Vis.})

@String(ECCV= {Eur. Conf. Comput. Vis.})

@String(ICLR = {Int. Conf. Learn. Represent.})

@String(AAAI = {AAAI})

@String(CVPR  = {CVPR})

@String(ICCV  = {ICCV})

@String(ECCV  = {ECCV})

@String(ICLR  = {ICLR})

@inproceedings{met3r,
    title = {MEt3R: Measuring Multi-View Consistency in Generated Images},
    author = {Asim, Mohammad and Wewer, Christopher and Wimmer, Thomas and Schiele, Bernt and Lenssen, Jan Eric},
    booktitle = {IEEE/CVF Computer Vision and Pattern Recognition ({CVPR})},
    year = {2025},
}

@InProceedings{dust3r,
    author    = {Wang, Shuzhe and Leroy, Vincent and Cabon, Yohann and Chidlovskii, Boris and Revaud, Jerome},
    title     = {DUSt3R: Geometric 3D Vision Made Easy},
    booktitle = {Proceedings of the IEEE/CVF Conference on Computer Vision and Pattern Recognition (CVPR)},
    month     = {June},
    year      = {2024},
    pages     = {20697-20709}
}

@inproceedings{vggt,
  title={VGGT: Visual Geometry Grounded Transformer},
  author={Wang, Jianyuan and Chen, Minghao and Karaev, Nikita and Vedaldi, Andrea and Rupprecht, Christian and Novotny, David},
  booktitle={Proceedings of the IEEE/CVF Conference on Computer Vision and Pattern Recognition},
  month={June},
  year={2025},
  pages={5294--5306}
}

@inproceedings{cut3r,
    author = {Wang, Qianqian and Zhang, Yifei and Holynski, Aleksander and Efros, Alexei A. and Kanazawa, Angjoo},
    title = {Continuous 3D Perception Model with Persistent State},
    booktitle={Proceedings of the IEEE/CVF Conference on Computer Vision and Pattern Recognition (CVPR)},
    month={June},
    year={2025},
    pages={10510--10522}
}

@inproceedings{pi3,
  title={$\pi^{3}$: Permutation-Equivariant Visual Geometry Learning},
  author={Wang, Yifan and Zhou, Jianjun and Zhu, Haoyi and Chang, Wenzheng and Zhou, Yang and Li, Zizun and Chen, Junyi and Pang, Jiangmiao and Shen, Chunhua and He, Tong},
  booktitle={International Conference on Learning Representations},
  year={2026}
}

@misc{mast3r,
      title={Grounding Image Matching in 3D with MASt3R}, 
      author={Vincent Leroy and Yohann Cabon and Jerome Revaud},
      booktitle = {ECCV},
      year = {2024}
}

@article{3dgs,
  title={3D Gaussian Splatting for Real-Time Radiance Field Rendering},
  author={Kerbl, Bernhard and Kopanas, Georgios and Leimk{\"u}hler, Thomas and Drettakis, George},
  journal={ACM Transactions on Graphics},
  volume={42},
  number={4},
  pages={1--14},
  year={2023}
}

@inproceedings{argus,
  title={Beyond the Frame: Generating 360{\textdegree} Panoramic Videos from Perspective Videos},
  author={Luo, Rundong and Wallingford, Matthew and Farhadi, Ali and Snavely, Noah and Ma, Wei-Chiu},
  booktitle={Proceedings of the IEEE/CVF International Conference on Computer Vision (ICCV)},
  year={2025}
}

@inproceedings{360dvd,
  title={360DVD: Controllable Panorama Video Generation with 360-Degree Video Diffusion Model},
  author={Wang, Qian and Li, Weiqi and Mou, Chong and Cheng, Xinhua and Zhang, Jian},
  booktitle={Proceedings of the IEEE/CVF Conference on Computer Vision and Pattern Recognition (CVPR)},
  month={June},
  year={2024},
  pages={6913--6923}
}

@article{gsplat,
  title={gsplat: An open-source library for Gaussian splatting},
  author={Ye, Vickie and Li, Ruilong and Kerr, Justin and Turkulainen, Matias and Yi, Brent and Pan, Zhuoyang and Seiskari, Otto and Ye, Jianbo and Hu, Jeffrey and Tancik, Matthew and Angjoo Kanazawa},
  journal={Journal of Machine Learning Research},
  volume={26},
  number={34},
  pages={1--17},
  year={2025}
}

@inproceedings{wonderjourney,
    title={Wonderjourney: Going from Anywhere to Everywhere},
    author={Hong-Xing Yu and Haoyi Duan and Junhwa Hur and Kyle Sargent and Michael Rubinstein and William T. Freeman and Forrester Cole and Deqing Sun and Noah Snavely and Jiajun Wu and Charles Herrmann},
    booktitle={CVPR},
    year={2024}
}

@inproceedings{vidpanos,
      title={VidPanos: Generative Panoramic Videos from Casual Panning Videos}, 
      author={Jingwei Ma and Erika Lu and Roni Paiss and Shiran Zada and Aleksander Holynski and Tali Dekel and Brian Curless and Michael Rubinstein and Forrester Cole},
      booktitle={SIGGRAPH Asia 2024 Conference Papers},
      month={December},
      year={2024}
}

@inproceedings{imagine360,
  title={Imagine360: Immersive 360 Video Generation from Perspective Anchor},
  author={Tan, Jing and Yang, Shuai and Wu, Tong and He, Jingwen and Guo, Yuwei and Liu, Ziwei and Lin, Dahua},
  booktitle={Advances in Neural Information Processing Systems (NeurIPS)},
  year={2025}
}

@article{panowan,
  title={PanoWan: Lifting Diffusion Video Generation Models to 360{\textdegree} with Latitude/Longitude-aware Mechanisms},
  author={Xia, Yifei and Weng, Shuchen and Yang, Siqi and Liu, Jingqi and Zhu, Chengxuan and Teng, Minggui and Jia, Zijian and Jiang, Han and Shi, Boxin},
  journal={arXiv preprint arXiv:2505.22016},
  year={2025}
}

@inproceedings{viewpoint,
  title={ViewPoint: Panoramic Video Generation with Pretrained Diffusion Models},
  author={Fang, Zixun and Zhu, Kai and Liu, Zhiheng and Liu, Yu and Zhai, Wei and Cao, Yang and Zha, Zheng-Jun},
  booktitle={Advances in Neural Information Processing Systems (NeurIPS)},
  year={2025}
}

@article{360anything,
  title={360Anything: Geometry-Free Lifting of Images and Videos to 360{\textdegree}},
  author={Wu, Ziyi and Watson, Daniel and Tagliasacchi, Andrea and Fleet, David J. and Brubaker, Marcus A. and Saxena, Saurabh},
  journal={arXiv preprint arXiv:2601.16192},
  year={2026}
}

@inproceedings{cubecomposer,
    title={CubeComposer: Spatio-Temporal Autoregressive 4K 360° Video Generation from Perspective Video},
    author={Li, Lingen and Wang, Guangzhi and Li, Xiaoyu and Zhang, Zhaoyang and Dou, Qi and Gu, Jinwei and Xue, Tianfan and Shan, Ying},
    booktitle={Proceedings of the IEEE/CVF Conference on Computer Vision and Pattern Recognition (CVPR)},
    year={2026}
}

@inproceedings{dit360,
  title={DiT360: High-Fidelity Panoramic Image Generation via Hybrid Training},
  author={Feng, Haoran and Zhang, Dizhe and Li, Xiangtai and Du, Bo and Qi, Lu},
  booktitle={Proceedings of the IEEE/CVF Conference on Computer Vision and Pattern Recognition (CVPR)},
  year={2026}
}

@article{panodiffusion,
  title={PanoDiffusion: 360-degree Panorama Outpainting via Diffusion},
  author={Wu, Tianhao and Zheng, Chuanxia and Cham, Tat-Jen},
  journal={arXiv preprint arXiv:2307.03177},
  year={2023}
}

@article{diffusion360,
  title={Diffusion360: Seamless 360 Degree Panoramic Image Generation based on Diffusion Models},
  author={Feng, Mengyang and Liu, Jinlin and Cui, Miaomiao and Xie, Xuansong},
  journal={arXiv preprint arXiv:2311.13141},
  year={2023}
}

@article{diffpano,
      title={DiffPano: Scalable and Consistent Text to Panorama Generation with Spherical Epipolar-Aware Diffusion},
      author={Weicai Ye and Chenhao Ji and Zheng Chen and Junyao Gao and Xiaoshui Huang and Song-Hai Zhang and Wanli Ouyang and Tong He and Cairong Zhao and Guofeng Zhang},
      booktitle={NeurIPS},
      year={2024},
  }

@inproceedings{panfusion,
  title={Taming Stable Diffusion for Text to 360◦ Panorama Image Generation},
  author={Zhang, Cheng and Wu, Qianyi and Cruz Gambardella, Camilo and Huang, Xiaoshui and Phung, Dinh and Ouyang, Wanli and Cai, Jianfei},
  booktitle={Proceedings of the IEEE/CVF Conference on Computer Vision and Pattern Recognition},
  year={2024}
}

@inproceedings{panofree,
  title={PanoFree: Tuning-Free Holistic Multi-view Image Generation with Cross-view Self-Guidance},
  author={Liu, Aoming and Li, Zhong and Chen, Zhang and Li, Nannan and Xu, Yi and Plummer, Bryan A},
  booktitle={European Conference on Computer Vision (ECCV)},
  year={2024},
}

@inproceedings{
spnd,
title={Spherical-Nested Diffusion Model for Panoramic Image Outpainting},
author={Xiancheng Sun and Senmao Ma and Shengxi Li and Mai Xu and Jingyuan Xia and Lai Jiang and Xin Deng and Jiali Wang},
booktitle={Forty-second International Conference on Machine Learning},
year={2025},
}

@InProceedings{panodecouple,
      title={Panorama Generation From NFoV Image Done Right},
      author={Zheng, Dian and Zhang, Cheng and Wu, Xiao-Ming and Li, Cao and Lv, Chengfei and Hu, Jian-Fang and Zheng, Wei-Shi},
      booktitle={Proceedings of the IEEE/CVF Conference on Computer Vision and Pattern Recognition},
      year={2025}
}

@inproceedings{dynamicscaler,
    title={DynamicScaler: Seamless and Scalable Video Generation for Panoramic Scenes},
    author={Liu, Jinxiu and Lin, Shaoheng and Li, Yinxiao and Yang, Ming-Hsuan},
    booktitle={Proceedings of the Computer Vision and Pattern Recognition Conference (CVPR)},
    year={2025},
}

@article{jopano,
  title={JoPano: Unified Panorama Generation via Joint Modeling},
  author={Feng, Wancheng and An, Chen and He, Zhenliang and Kan, Meina and Shan, Shiguang and Wang, Lukun},
  journal={arXiv:2512.06885},
  year={2025}
}

@article{spherediff,
  title={SphereDiff: Tuning-free Omnidirectional Panoramic Image and Video Generation via Spherical Latent Representation},
  author={Park, Minho and Kang, Taewoong and Yun, Jooyeol and Hwang, Sungwon and Choo, Jaegul},
  journal={arXiv preprint arXiv:2504.14396},
  year={2025}
}

@misc{cubediff,
      title={CubeDiff: Repurposing Diffusion-Based Image Models for Panorama Generation}, 
      author={Nikolai Kalischek and Michael Oechsle and Fabian Manhardt and Philipp Henzler and Konrad Schindler and Federico Tombari},
      year={2025},
      eprint={2501.17162},
      archivePrefix={arXiv},
      primaryClass={cs.CV},
      url={https://arxiv.org/abs/2501.17162}, 
}

@inproceedings{dreamcube,
    title     = {{DreamCube: RGB-D Panorama Generation via Multi-plane Synchronization}},
    author    = {Huang, Yukun and Zhou, Yanning and Wang, Jianan and Huang, Kaiyi and Liu, Xihui},
    booktitle = {Proceedings of the IEEE/CVF International Conference on Computer Vision (ICCV)},
    month     = {October},
    year      = {2025},
    pages     = {24922-24932}
}

@inproceedings{personalizeanything,
  title={Personalize anything for free with diffusion transformer},
  author={Feng, Haoran and Huang, Zehuan and Li, Lin and Sheng, Lu},
  booktitle={Proceedings of the AAAI Conference on Artificial Intelligence},
  volume={40},
  number={5},
  pages={3921--3929},
  year={2026}
}

@article{worldmirror,
  title={WorldMirror: Universal 3D World Reconstruction with Any-Prior Prompting},
  author={Liu, Yifan and Min, Zhiyuan and Wang, Zhenwei and Wu, Junta and Wang, Tengfei and Yuan, Yixuan and Luo, Yawei and Guo, Chunchao},
  journal={arXiv preprint arXiv:2510.10726},
  year={2025}
}

@inproceedings{pixelsplat,
  title={pixelSplat: 3D Gaussian Splats from Image Pairs for Scalable Generalizable 3D Reconstruction},
  author={Charatan, David and Li, Sizhe Lester and Tagliasacchi, Andrea and Sitzmann, Vincent},
  booktitle={Proceedings of the IEEE/CVF Conference on Computer Vision and Pattern Recognition (CVPR)},
  month={June},
  year={2024},
}

@inproceedings{mvsplat,
  title={MVSplat: Efficient 3D Gaussian Splatting from Sparse Multi-View Images},
  author={Chen, Yuedong and Xu, Haofei and Zheng, Chuanxia and Zhuang, Bohan and Pollefeys, Marc and Geiger, Andreas and Cham, Tat-Jen and Cai, Jianfei},
  booktitle={European Conference on Computer Vision (ECCV)},
  year={2024},
  pages={370--386}
}

@article{anysplat,
  title={AnySplat: Feed-forward 3D Gaussian Splatting from Unconstrained Views},
  author={Jiang, Lihan and Mao, Yucheng and Xu, Linning and Lu, Tao and Ren, Kerui and Jin, Yichen and Xu, Xudong and Yu, Mulin and Pang, Jiangmiao and Zhao, Feng and Lin, Dahua and Dai, Bo},
  journal={ACM Transactions on Graphics},
  volume={44},
  number={6},
  pages={1--16},
  year={2025}
}

@article{noposplat,
      title   = {No Pose, No Problem: Surprisingly Simple 3D Gaussian Splats from Sparse Unposed Images},
      author  = {Ye, Botao and Liu, Sifei and Xu, Haofei and Xueting, Li and Pollefeys, Marc and Yang, Ming-Hsuan and Songyou, Peng},
      journal = {arXiv preprint arXiv:2410.24207},
      year    = {2024}
    }

@inproceedings{yonosplat,
  title     = {YoNoSplat: You Only Need One Model for Feedforward 3D Gaussian Splatting},
  author    = {Ye, Botao and Chen, Boqi and Xu, Haofei and Barath, Daniel and Pollefeys, Marc},
  booktitle = {International Conference on Learning Representations (ICLR)},
  year      = {2026}
}

@inproceedings{lpips,
  title={The Unreasonable Effectiveness of Deep Features as a Perceptual Metric},
  author={Zhang, Richard and Isola, Phillip and Efros, Alexei A. and Shechtman, Eli and Wang, Oliver},
  booktitle={Proceedings of the IEEE Conference on Computer Vision and Pattern Recognition},
  pages={586--595},
  year={2018}
}

@article{pager,
  title   = {Unified Panoramic Geometry Estimation via Multi-View Foundation Models},
  author  = {Bozic, Vukasin and Slavkovic, Isidora and Narnhofer, Dominik and
             Metzger, Nando and Rozumny, Denis and Schindler, Konrad and
             Kalischek, Nikolai},
  journal = {arXiv preprint arXiv:2605.26368},
  year    = {2026}
}

@inproceedings{vbench,
  title={VBench: Comprehensive Benchmark Suite for Video Generative Models},
  author={Huang, Ziqi and He, Yinan and Yu, Jiashuo and Zhang, Fan and Si, Chenyang and Jiang, Yuming and Zhang, Yuanhan and Wu, Tianxing and Jin, Qingyang and Chanpaisit, Nattapol and Wang, Yaohui and Chen, Xinyuan and Wang, Limin and Lin, Dahua and Qiao, Yu and Liu, Ziwei},
  booktitle={Proceedings of the IEEE/CVF Conference on Computer Vision and Pattern Recognition (CVPR)},
  month={June},
  year={2024},
}

@article{telea,
  title={An Image Inpainting Technique Based on the Fast Marching Method},
  author={Telea, Alexandru},
  journal={Journal of Graphics Tools},
  volume={9},
  number={1},
  pages={23--34},
  year={2004}
}

@article{qwen3vl,
      title={Qwen3-VL Technical Report}, 
      author={Shuai Bai and Yuxuan Cai and Ruizhe Chen and Keqin Chen and Xionghui Chen and Zesen Cheng and Lianghao Deng and Wei Ding and Chang Gao and Chunjiang Ge and Wenbin Ge and Zhifang Guo and Qidong Huang and Jie Huang and Fei Huang and Binyuan Hui and Shutong Jiang and Zhaohai Li and Mingsheng Li and Mei Li and Kaixin Li and Zicheng Lin and Junyang Lin and Xuejing Liu and Jiawei Liu and Chenglong Liu and Yang Liu and Dayiheng Liu and Shixuan Liu and Dunjie Lu and Ruilin Luo and Chenxu Lv and Rui Men and Lingchen Meng and Xuancheng Ren and Xingzhang Ren and Sibo Song and Yuchong Sun and Jun Tang and Jianhong Tu and Jianqiang Wan and Peng Wang and Pengfei Wang and Qiuyue Wang and Yuxuan Wang and Tianbao Xie and Yiheng Xu and Haiyang Xu and Jin Xu and Zhibo Yang and Mingkun Yang and Jianxin Yang and An Yang and Bowen Yu and Fei Zhang and Hang Zhang and Xi Zhang and Bo Zheng and Humen Zhong and Jingren Zhou and Fan Zhou and Jing Zhou and Yuanzhi Zhu and Ke Zhu},
	  journal={arXiv preprint arXiv:2511.21631},
      year={2025}
}

@inproceedings{wonderworld,
    title={WonderWorld: Interactive 3D Scene Generation from a Single Image},
    author={Hong-Xing Yu and Haoyi Duan and Charles Herrmann and William T. Freeman and Jiajun Wu},
    booktitle={CVPR},
    year={2025}
}

@article{neoverse,
  author    = {Yang, Yuxue and Fan, Lue and Shi, Ziqi and Peng, Junran and Wang, Feng and Zhang, Zhaoxiang},
  title     = {NeoVerse: Enhancing 4D World Model with in-the-wild Monocular Videos},
  journal   = {arXiv preprint arXiv:2601.00393},
  year      = {2026},
}

@misc{srender,
      title={Efficient Camera-Controlled Video Generation of Static Scenes via Sparse Diffusion and 3D Rendering}, 
      author={Jieying Chen and Jeffrey Hu and Joan Lasenby and Ayush Tewari},
      year={2026},
      eprint={2601.09697},
      archivePrefix={arXiv},
      primaryClass={cs.CV},
      url={https://arxiv.org/abs/2601.09697}, 
}

@inproceedings{gen3c,
        title={GEN3C: 3D-Informed World-Consistent Video Generation with Precise Camera Control},
        author={Ren, Xuanchi and Shen, Tianchang and Huang, Jiahui and Ling, Huan and 
            Lu, Yifan and Nimier-David, Merlin and M\"uller, Thomas and Keller, Alexander and 
            Fidler, Sanja and Gao, Jun},
        booktitle={Proceedings of the IEEE/CVF Conference on Computer Vision and Pattern Recognition},
        year={2025}
    }

@InProceedings{trajectorycrafter,
    author    = {Yu, Mark and Hu, Wenbo and Xing, Jinbo and Shan, Ying},
    title     = {TrajectoryCrafter: Redirecting Camera Trajectory for Monocular Videos via Diffusion Models},
    booktitle = {Proceedings of the IEEE/CVF International Conference on Computer Vision (ICCV)},
    month     = {October},
    year      = {2025},
    pages     = {100-111}
}

@article{viewcrafter,
    title={ViewCrafter: Taming Video Diffusion Models for High-fidelity Novel View Synthesis},
    author={Yu, Wangbo and Xing, Jinbo and Yuan, Li and Hu, Wenbo and Li, Xiaoyu and Huang, Zhipeng and Gao, Xiangjun and Wong, Tien-Tsin and Shan, Ying and Tian, Yonghong},
    journal={arXiv preprint arXiv:2409.02048},
    year={2024}
  }

@misc{camtrol,
      title={Training-free Camera Control for Video Generation}, 
      author={Chen Hou and Zhibo Chen},
      year={2025},
      eprint={2406.10126},
      archivePrefix={arXiv},
      primaryClass={cs.CV},
      url={https://arxiv.org/abs/2406.10126}, 
}
